# Augmented Random Search for Quadcopter Control: An alternative to Reinforcement Learning

**Ashutosh Kumar Tiwari**
MEMBER TECHNICAL STAFF, ORACLE INC, BANGALORE, INDIA
E-mail: reachatashutosh@gmail.com

**Sandeep Varma Nadimpalli**
Assistant Professor (INFORMATION SCIENCE AND ENGINEERING), BMSCE, BANGALORE, INDIA
E-mail: sandeepvarma.ise@bmsce.ac.in

*Abstract*—Model-based reinforcement learning strategies are believed to exhibit more significant sample complexity than model-free strategies to control dynamical systems, such as quadcopters. This belief that Model-based strategies that involve the use of well-trained neural networks for making such high-level decisions always give better performance can be dispelled by making use of Model-free policy search methods. This paper proposes the use of a model-free random searching strategy, called Augmented Random Search (ARS), which is a better and faster approach of linear policy training for continuous control tasks like controlling a Quadcopter's flight. The method achieves state-of-the-art accuracy by eliminating the use of too much data for the training of neural networks that are present in the previous approaches to the task of Quadcopter control. The paper also highlights the performance results of the searching strategy used for this task in a strategically designed task environment with the help of simulations. Reward collection performance over 1000 episodes and agent's behavior in flight for augmented random search is compared with that of the behavior for reinforcement learning state-of- the-art algorithm, called Deep Deterministic policy gradient(DDPG) Our simulations and results manifest that a high variability in performance is observed in commonly used strategies for sample efficiency of such tasks but the built policy network of ARS-Quad can react relatively accurately to step response providing a better performing alternative to reinforcement learning strategies.

## I. INTRODUCTION

The concept of model-based Reinforcement learning (RL) [1-3]provides a competitive approach to control dynamical frameworks. In such strategies, the models of the framework's elements play a key role in control and have effectively delivered RL agents that outperform human players in many such continuous control problems. Although these outcomes are imposing, methods that are not based on models (model-free) have not still been advantageously utilized to control such physical aerial frameworks . It is a common belief that that model-based strategies provide better sample complexity than the model-free searching strategies that explore the space of actions for autonomous control tasks. Such beliefs are dispelled by the use of the proposed simple searching strategy which takes advantage of eliminating the use of excessive data to train neural networks and matches the state-of-the-art accuracy for such tasks by a random search in the policy space.

Continuous control tasks are majorly addressed using state-of- the-art RL methods and strategies on agents for good and sample efficient results. Quadcopter control. Several previous works by authors suggest the use of RL with Least Square policy Iteration(LSPI) [4] to learn optimum control policies for quadcopter control. Few works also indicate the use of state- of-the-art RL algorithms like DDPG, TRPO and PPO to perform such tasks [5]. But, there exist few elements precluding the reception of RL methods for controlling these physical frameworks: the strategies demand a lot of information to accomplish sensible execution, the regularly expanding combination of RL techniques tend to make it hard to pick what is the best technique for a particular undertaking and numerous applicant calculations are difficult to utilize.

Tragically the momentum pattern in RL based research has put these hindrances inconsistent with one another. In this mission to discover strategies that test productive (i.e. techniques that need very little information), the general pattern has been to develop progressively convoluted solutions to such continuous control problems [8,9]. This has resulted in a reproducibility crisis. Recent examinations for such methods exhibit that numerous RL based methods are not sturdy to changes in hyperparameters, randomized seed values or even various usage of the identical algorithm. Algorithms that involve such daintiness cannot be coordinated into continuous mission-critical control [10] frameworks without much improvement and robustification. Also, it's common to measure and evaluate the performance of it by applying it to simulated continuous control problems over a small number of trials.

The scope of this work is to use the simplest model-free RL method involving random searching that can give encouraging results in learning-based tasks with continuous state space and action space [11, 12]. This work shows that an aerial system framework i.e Quadcopter [13-15] can be completely controlled utilizing a simple random searching technique called augmented random search, without taking into consideration of excessive data for training of agents on reinforcement learning algorithms. The strategy used is a straightforward mapping of a state of the agent (quadcopter) thrust so there are only a few assumptions made for the physical framework of the copter.

Also, policy-based learning [16,17] on any aerial system is frequently demonstrated in autonomous control literature. This work has shown that natural policy gradients can be used to train linear policies to obtain high-performance solutions to continuous control problems like a quadcopter flying. The simulation shows that a more dynamically characterized motion can be accomplished through a simple augmented random search technique. This work combines the ideas of background research done on quadcopter control using reinforcement learning techniques and a high performance simple random searching technique called augmented random search[18]. The main contribution of this research is done by introducing an alternative, better and faster learning algorithm to train a quadcopter to fly without using model-based deterministic on-policy methods of reinforcement learning. This works on a simple derivative-free optimization strategy which can surpass the results of using a zero-bias, zero-variance samples used in reinforcement learning strategies.

Our work contributes to the task of controlling the quadcopter's flight in the following ways,

1. Building the physical simulator structure and framework of the agent which will perform the task of flying (Quadflytask).
2. Strategically designing the task environment that defines the goal and provides feedback to the agent
3. Using the augmented random search algorithm ARS- Quad for the online normalization of states and updating rewards to help the agent learn.

## II. RELATED WORK

The presented approach uses a derivative-free policy optimization method[20] whose early interpretations have been proposed for facilitating future research on RL. *Salimans et al.* [3] accomplished a similar strategy called the Evolution strategies which showed that this method could easily be used for training policies faster than alternative methods of performing such tasks. The algorithm proposed by *Saliman et al. [3]* involved several complicated algorithmic details in spite of being simpler than previously proposed methods. Another simplified approach for the derivative-free policy and model-free RL was proposed by *Rajeshwaran et al.* [6] which showed the use of natural policy gradients for continuous control tasks. Apart from this, various interpretations of quadcopter control have also come to the forefront by the use of deterministic ways of policy optimization using a natural gradient descent.

Extending the idea of using deterministic policies instead of stochastic policies to control the quadcopter the present work has been put into picture. Controlling the quadcopter using stochastic policy [19] can lead to unpredictable performance and does not serve the purpose. Also using deterministic policy optimization over stochastic policy optimization gives the privileges of having lower variance of value estimates from on-policy samples. Deterministic policy gradient methods [20] requires a good strategy for exploration to explore the state space, unlike the stochastic strategy based policy gradient.

Any reinforcement learning problem requires searching for policies to control dynamical frameworks or systems that can maximize an average reward given to the agent by the system. Such problems can be stated as,

$$\max \mathrm{E}\, \xi[\, \mathrm{r}\,(\, \pi_\theta\,,\xi\,)] \qquad (1)$$

In (1), $\theta$ characterizes a basic policy $\pi\theta$: $\mathbb{R}^n \to \mathbb{R}^m$. Also, the random variable $\xi$ shows the randomness and unplanned nature of the environment chosen to perform tasks i.e variable initial states and transitions which are stochastic in nature. For the same policy $\pi\theta$ on one direction or trajectory generated by the framework, value [ $(\pi\theta, \xi)$ ] is the reward gained by training the policy. Taking all this into consideration, the deep reinforcement learning techniques use stochastic policies for such problems but the ARS use for the flying task of quadcopter uses deterministic policies [21]. Random searching techniques have been the oldest and one of the simplest methods of carrying out optimization which is supposed to be derivative-free just by making use of approximation of finite-difference along chosen variable and random directions.

Augmented random search (ARS) [18,21] has proved to be the best breakthrough in this regard which relies on the various augmentations of basic random search to build on accomplished and proven heuristics that were earlier utilized in deep reinforcement learning techniques. ARS simplifies the problem of using deep neural networks and calculating the action value functions by the transformation of rankings from rewards and using these rankings to form the update steps for the quadcopter to learn the trajectory. It helps the agent to bin the action space to encourage exploration.Work done on quadcopter control depends on policies characterized by deep neural networks with virtual batch normalizations while using the ARS algorithm for the same task achieves leading-edge and competitive performance of quadcopter control with policies that are linear in nature.

## III. METHODOLOGY

The following section recounts the method that is employed in the present work for the training of the policy for a quadcopter. The rationality and viability of the method employed should be further examined and extended in the future.

### A. Physical Framework

The helicopter is an aerial vehicle that uses briskly spinning rotors which pushes air downwards to keep the helicopter skyward. This can be perceived as rotors, two in number, which are supposed to be coplanar both providing thrust in the same direction i.e. upwards but revolving in the opposite direction.

A quadrotor, usually called a quadcopter, is a form of helicopter which has four rotors equally spaced, generally disposed at the corners of a body, square in shape. Controlling a Quadcopter is essentially tedious to solve and an interesting problem. The dynamics of the quadcopter [22-26] are highly non-linear, taking into account the extensive aerodynamic effects. Also since quadcopters have only slight amounts of friction to intercept their trajectory of motion, so they have to facilitate their damping to halt the movement and remain steady.

This work uses a very straightforward and uncomplicated model for simulation purposes as for the purpose to be served one does not need to jump into every attribute of the quadcopter. We can disregard all the drag forces being acted upon on the quadcopter's body and utilize a simple and fairly elementary body model with the basic idea of having four thrust/propel forces acting on its body.

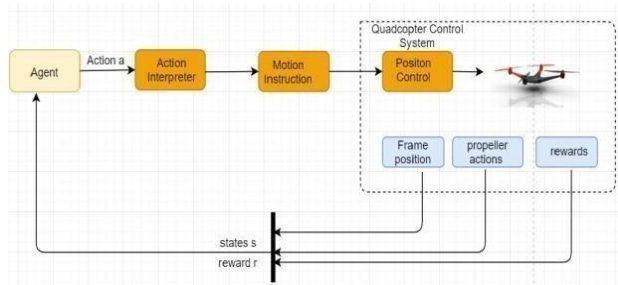

Fig.1. Agent (Quadcopter) reward mechanism

Fig. 1 illustrates the reward mechanism of the agent and the interconnection of its components that will help in simulation. It shows how the agent's control system interprets instructions from the environment to gain rewards. The agent's behavior is governed by a set of parameters (implemented as routines) which are follows,

- Earth to Body frame – It takes three inputs as the inputs of inertial axes to decide the earth-based origin at the launch location.
- Body to earth frame – Takes the same inputs to decide the center of gravity of the quadrotor aligned along the given frame.
- Simulation Function – Takes the inputs in the form of the initial position of the quadcopter, initial velocity and initial angular velocity with the default probable runtime of the engine.
- Propeller thrust – A routine that calculates the net thrust based on the velocity of the quadcopter.
- Propeller wind speed – A routine that calculates the propeller speed from the angular velocity and body velocity data.

The above functions written to build the robot model of simulation has been taken and modified to suit the purpose of this research work from the paper Reinforcement learning-based Quadcopter Control by Shayegan [14]. The virtual simulation of the robot model shows a very manageable model emphasizing that it is possible to accomplish quality performance even without putting in much effort to dummy any specifics of physics for the quadcopter dynamics.

### B. Task Environment

The research work evaluates the performance of ARS on the continuous control problem of making aerial frameworks and systems like quadcopters fly as tasks included in the environment created using the robot-like simulation model of the agent i.e the quadcopter. A task environment [10] is a set of states that the agent is trying to influence through the choice of its actions. So the structure of the task environment depends broadly on the signals which are relevant to the task, and how they interact. The boundary between the agent and the task environment can be chosen at different and separate places and for different purposes. In particular, the boundary of agent-environment is to determined when one has particular areas selected, specific actions and rewards, thereby has established a specific task of decision making for the interest. The task environment performs the function of defining a goal and provides feedback to the agent.

Algorithm I. Task Environment for defining the goal and providing feedback to the agent

**1**: **Initialize** a task object O which will act as the agent A exploring the linear policy $\pi\theta$.
**2: Parameters :**

( , $y_i$ , $z_i$ ) , as the initial position of quadcopter for i ∈ { 0, 1…,N}, say (0. , 0. , 0.), and the Euler angles ( $\Phi$, $\theta$ , $\Psi$)

($v_x$, $v_y$, $v_z$) , as the initial velocity of quadcopter {x,y,z} in the action space S

($\omega x$ , $\omega y$ , $\omega z$) , as the initial angular velocity for each

Euler angle ( , $\theta$ , $\Psi$)

T , as the time limit for each episode or runtime

$(t_x, t_{yi}, t_{zi})$, as the target /goal position for the agent

**3**: **while** an episode persists **do**

**4**: For the task object O, use current position$(x_i, y_i, z_i)$ in the action space S to return the reward R collected by exploration under a hyperparameter $\partial$ which depends on the count of possible directions of movement of quadcopter

$$R= R - [sum ( | (x_i, y_i, z_i) - (t_{xi}, t_{yi}, t_{zi})|)] \quad (2)$$

**5** : The Agent A, quadcopter performs a series of actions ( $a_1, a_2,...a_N$ ) in the action space S which is used to move to next state and get the next set of rewards by using (2)

**6**: **end** of episode

**7**: **reset** the environment and state space P , for a new episode in order to perform exploration and exploitation

**8**: For the task object , separately design a takeoff *TO(v)* routine and its corresponding reward function *TOR (v)* for propeller speed v,

*TOR(v)* : Use the current position $(xi, yi, zi)$ to return the reward,

If ( $|(t_{zi}) - (zi)|$ ) < ( $|(t_{zi}) - (zi-1)|$ ) :

  R = R – { *penalizing with suitable random value* }
Else
  R = R + { *rewarding with suitable random value* }
  where ( $xi-1, yi-1, zi-1$) is the last/previous position of
   the quadcopter

If ( ) >= ( $i-1$) :
  R = R + { *rewarding with larger random suitable value for crossing the target height* }

Update : $(zi-1) = (zi)$
*TO(v)* : for iteration i=1 to k in series of actions $(a1, a2,...aN)$ ,
  R = R + *TOR(v)* ,
Move to the next state and get the next set of rewards

---

The code used to simulate the quadcopter and its task environment is written for this research work to ensure that it is both accurate numerically and performance wise and also stable. Since the simulation task is also written in python, the time required for the computation for the dynamics to be integrated is far less than otherwise using deep neural networks for the same purpose in case of quadcopter control using deep reinforcement learning techniques.

*C. Augmented Random Search Algorithm for Quadcopter*

The Augmented random search (ARS) algorithm relies on the augmentations of the basic version of the random search algorithm that is in turn built on accomplished and proven heuristics which is utilized in deep reinforcement learning strategies and techniques [28-31]. The prime problem of policy search based on augmentation can be formulated and analyzed as being a continuous problem of searching i.e a continuous search problem. The various operations in the task of flying the quadcopter are performed in a definite order and the search space is also continuous, which helps to make suré that the process of searching is diverse and simple. The main idea of quadcopter control using augmented random search is to search from the parameter space the best possible policy $\pi\theta$. In a more detailed form, it analyzes a series of possible directions in the parameter space to collect appropriate rewards and keep on optimizing the step along each such possible direction to form the best possible policy.

To optimize the steps in each possible direction to reach the goal of forming the final best policy, the algorithm performs the update of each perturbation direction by calculating the difference of the rewards $r$ $(\pi_i, j, +)$ and $r (\pi_i, j, -)$. This routine estimates the step to move in a specific direction. Also,with this the updating process has been improved by eradication of update steps calculations for the directions that result in the least improvement of reward. The use of such a technique can assure that update steps employed are averaged over the directions that have gained high and quality rewards [32,33].

Algorithm II. Implementation of augmented random search for the quadcopter (ARS-Quad)

**1**: **Hyperparameters**: size of step, count of directions taken into consideration per iteration $N$, exploration noise standard deviation  , count of the best performing directions for use $\alpha$

**2**: **Initialize** matrix M with all zeros , $M0= 0 \epsilon R^{m \times n}$, i = 0

**3**: **repeat while**

**4**: **Sample** $\partial 1, \partial 2, .... , \partial N$ in $R^{m \times n}$ with independently and identically distributed Standard normal Inputs

**5**: Collecting 2N outputs with the use of 2N policies

  $\pi i, j, + (x)$= (M i + $\mu \partial k$) $diag(\Sigma_i)^{-1/2}$ ( $x - \mu_i$)
  $\pi i, j, - (x)$= (M i - $\mu \partial k$) $diag(\Sigma_i)^{-1/2}$ ( $x - \mu_i$)
  for $k \in \{1,2, ...., N\}$

**6**: Collect each output $\pi_i, j$ into a separate tracer routine *t(x)* initialized as lists of states, rewards , actions and traces

**7**: Tracer routine *t(x)* will keep track of the states, rewards and actions with traces per episode
**8**: Sorting of directions $\partial_k$ by { $r(\pi_i, j, +)$, $r(\pi_i, j, -)$}, denote by $\partial_k$ the k-th largest direction of perturbation and corresponding policies by the terms in the bracket
**9**: Proceed with update step :

$$M_{i+1} = M_i + \frac{\beta}{\alpha \sigma_R} \sum_{k=1}^{\alpha} [r(\pi_i, j, +) - r(\pi_i, j, -)] \partial_k$$

where $\sigma_R$ is the SD of the $2\alpha$ rewards used for update
**10**: $i = i + 1$
**11: end loop**

The three essential steps for the quadcopter to learn its trajectory of motion is the following:

1. Normalization of states: The state normalization process is an important step in such regression similar tasks because it fortifies that the policies put equal weights on the different and in-process state components. This helps in balancing the control gain by making it larger for the smaller state coordinates and leading to smaller gains concerning larger state coordinates.
2. Scaling using standard deviation: Searching randomly in the policy space can result in large differences in the reward gained as the training of the policies progresses, as a result, it updates steps vary in steps and it becomes difficult to choose a fixed step size. Thus the rewards are transformed into rankings to compute the update step.
3. Using top-performing actions or directions with the best average rewards for the update step

The ARS algorithm used here with slight modification is taken and modified to our needs for the quadcopter flying task from the paper of *Horia mania et al.* [18]

## IV. RESULT AND DISCUSSION

The following section recounts the methods to evaluate the performance of the algorithm for the task of flying a quadcopter. The section also presents a comparative study of the presented algorithm with another model-free algorithmic approach called Deep DPG (Deterministic policy gradient) [9] which can learn policies competitively for continuous control tasks and makes use of deep reinforcement learning techniques, implying the utilization of deep Neural Nets.

### A. Experimental Setup

The implementation of the presented version of the algorithm was done using python3.6 and the various visualizations were developed using the python library matplotlib[36] and the open-source graphing library plotly [37]. The task environment is setup using the Algorithm I. The agent goal and feedback mechanism is dependent on the task environment. The environment defines the initial values of position(init_pose), velocities (init_velocities), angular velocities(init_ang_vel), time limit of each episode(runtime) and the target position(tar_pos). The task environment is responsible for the agent's reward collection and thus the authors implement step response functions, takeoff() and get_reward_takeoff() which are used to perform the action to move to a next state. After the reward for an ended episode is collected the agent parameters are reset using the reset() routine. The task environment is primarily responsible to initialize a task to be performed by the agent in order to get rewards.

The simulating agent absorbs the values of position, velocities and runtime initialized in the task environment already setup to perform actions and provide a step response in return. For simulations, the agent's actions, reward functions, and step rewards are implemented using routines written in python. The driving algorithm for the agent's behavior starts with setting up of task-specific hyperparameters consisting of number of steps (set to 200), episode length (set to 1000), learning rate (set to 0.01), number of directions (set to 16), number of best directions (set to 4) for specific random seed and noise values.

The policy where the agent performs exploration is then implemented. It is a function that takes states of the environment as inputs and returns the actions in order to help the agent learn the task. The agent performs exploration over a space of such policies using the explore() routine and converges to the one that returns the best actions or actions with the greatest positive reward. The agent performs policy exploration on one specific direction over one episode and the rewards are collected in a shared noise data file using a tracer() routine. The shared file is used to perform visualizations using Matplotlib and plotly.

As the agent selectively chooses the best step response by exploration, the space of policies over an episode is updated the reward obtained after each update is plotted using matplotlib and plotly. The degree to which an agent learns and how fastly depends on the number of positive and negative rewards collected over an episode. Therefore, in order to compare the behavior of the two agents, ARS-Quad which uses the random searching strategy for exploration and DDPG which uses the state-of-the-art deep deterministic policy gradient method for exploration, separate simulations were carried out over 1000 episodes for the same Quad_Fly task, results of which are presented in Fig. 4 and Fig. 5. The code was initially run on Intel Core i5-5200U@2.8GHz with 16 GB RAM and Nvidia Geforce Graphics where it gives optimal performance results. The final code was also run on Intel Pentium CPU 4405U @2.1GHz showing the feasibility of running the algorithm on slow-processors.

### B. Rewards for each episode of ARS-Quad

Further, to circumvent the bottleneck of computation of the perturbations or feasible direction of flying $\partial$, a shared noise data file which is responsible for storing independent normal entries, was created. This is done to ensure that the workers through the shared noise data file can interface the indices. *Moritz et al.* [7] have used such an approach for the implementation of evolution strategy in continuous control problems and also parallel to the approach of *Salimans et al.* [3] . In the presented work the random seed generators to facilitate the workers have also been set. To ensure a diverse sample efficiency the random seeds are kept distinct to each other. The training process was repeated 100 times with different and distinct random seeds and set of hyperparameters as discussed above to achieve a thorough searching of the policy space. The fixing of random seeds has been done by sampling uniformly from the interval [0,1000).

After the implementation of the task environment, it could provide standard functions that generated rewards to assess the performance and efficiency of policies, linearly trained using Algorithm II. The reward thresholds for various step responses were calculated and analyzed. Each episode associates a policy exploration and updates the step over a specific direction helping the agent to learn the task.

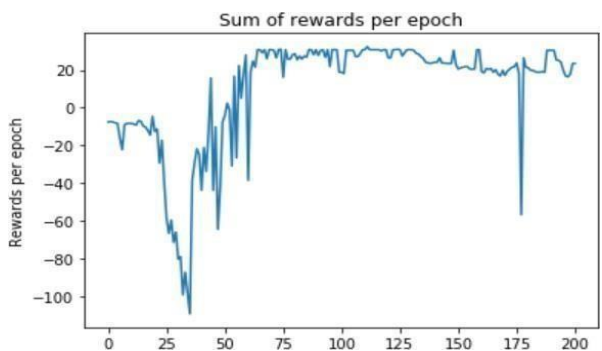

Fig.2. rewards vs sum of rewards per epoch

The Fig. 2 shows the rewards per epoch of training and the sum of average rewards per epoch. The increase and decrease of the average rewards per epoch shows that when the Agent A performs a set of actions ($a1, a2 \dots aN$) in the action space S, the rewards and penalties are applied on it in order to make the agent learn i.e help the quadcopter learn the correct trajectory of motion [34,35] by penalizing for each wrong takeoff/step and rewarding for each correct takeoff/step and movement towards the direction which provides the best rewards. It also fortifies the fact that as the epoch increases the perturbation directions that provide minimalistic reward improvement are penalized and removed.

Fig. 3 indicates the summary of the motion of quadcopter during the simulation of ARS-Quad , developed using the statistics which were saved in the shared noise data file while simulating the Quadfly task. In Fig. 3, the graph at the top left corner indicates how the position of the quadcopter evolved during the simulation. The three initial coordinates ( , $yi$ , $zi$) are set during the experiment to (0.,0.,10.0 ) to maintain the quadcopter at a starting height of 10 meters above the ground. The graph at the top right corner indicates the velocity of the quadcopter over the full simulation of the quadcopter. The graph at the bottom left corner indicates the velocity of the quadcopter wrt. each of the Euler angle in rad/sec, where the measure of Euler angles ( $\Phi, \theta , \Psi$) are an indicator of rotation of quadcopter over *x-*, *y-* , *z-* axes. The graph at the bottom right corner represents the ARS-Quad agent's choice of actions for each rotor per second during the simulation. The ARS-Quad agent selects an action for each rotor to set the revolutions per second on each of the four rotors to control the quadcopter. All the graphs represent the physical motion of the quadcopter over 1000 episodes.

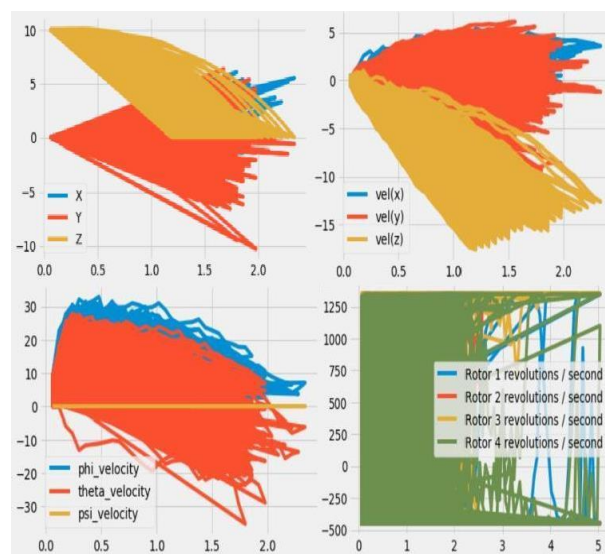

Fig.3. Summary of quadcopter motion during ARS-Quad

The following section of the study shows a compare and contrast survey between another model-free algorithmic approach called DDPG(Deep deterministic policy gradient)[9] and the simple linear policy searching strategy presented as ARS-Quad. The comparison is established by evaluating and analyzing plots of total rewards earned in each episode along with plots of how the quadcopter physically behaved for each of the two agents for the same Quadfly task. The ARS-Quad furnishes results that show optimum and substantial parallelism between the two different strategies of flying the quadcopter. While on one hand, DDPG uses complex neural net architecture for the task, ARS-Quad uses the Algorithm I to perform the same task giving competitive results.

RL methods are not expected to be subtle to choices of hyperparameters if one chooses to make actual use of them. However, DDPG is a model-free algorithmic approach that is very sensitive to choices of hyperparameters and hence makes it quite demanding and challenging to use in practice. In the experimental setup, the hyperparameters for the DDPG agent were carefully chosen depending on the neural net architecture, unlike ARS-Quad which used few and comparatively less complex hyperparameters being a simple random searching algorithmic approach. It was observed that the choice of hyperparameters does not heavily influence the success rate of ARS-quad and it is just affected by the choice of random seeds used for linear policy exploration. In other words, the ARS-quad is less sensitive to hyperparameter choice because its success rate when the hyperparameters are changed is equivalent to its success rate when independent trials are performed. Such experiments, therefore, gave ARS- Quad a competitive performance edge in the Quadflytask.

### C. DDPG vs ARS-Quad Agents

This section recounts of the behavior of both the agents in the simulated environment with details on how the agents physically behaved during the simulation.

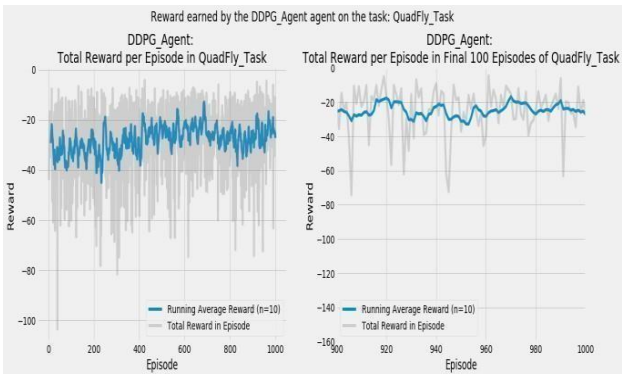

Fig.4. Reward earned by DDPG agent on the Quadfly task

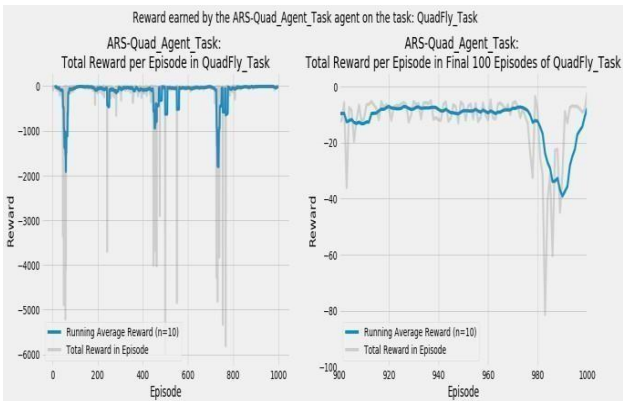

Fig.5. Reward earned by ARS-Quad agent on the Quadfly task

In Fig. 4, the graph on the left represents the total rewards earned in each episode of the Quadfly task by the DDPG agent along with the running average reward of the previous ten episodes (n=10). The graph on the right is a focused macro view to observe the rewards collected by the agent during the simulation of the last 100 episodes. The graph shows that in this period the DDPG agent has substantially and adequately learned the Quadfly task. During the 1000 episodes of the task, the agent earned a reward that varies within a stable range of roughly between -20 and -50. This fact indicates that the DDPG agent has been able to very well learn through rewards in 1000 episodes. Fig. 5 represents the same reward collection metrics for the same Quadfly task by the ARS-Quad agent. The graph on the right in Fig. 5, represents that the agent was able to continuously earn an average reward which is between -10 and -40 which is quite competitive to the reward collection of the DDPG agent. The graph show that towards the end of the 100 episodes of the simulation ARS-Quad agent's average reward dips down to around -40, but this dip is a small spot when comparing it to the prior results of the same simulation for 200, 400, 600 and 800 episodes. This indicates that we could expect the average reward to continue to converge to around -10 or below over subsequent episodes and shows that the agent is very well learning the task.

The following graphs in Fig.6 and Fig.7 represent the observation of how the quadcopter physically behaved during the 1000 episodes of simulation for both, the DDPG agent and the ARS-Quad agent on the same Quadfly task. The graphs in Fig. 6 and Fig. 7 indicates how well the DDPG and ARS-Quad agents were able to learn the goal of the Quadfly task. In the experimental setup, the copter begins each episode at (0., 0. ,10.0) as the initial position and is supposed to remain in this position indefinitely In the two figures, the graphs at the upper left corner represent the average position in *x, y, z* values of the quadcopter across all episodes at the first n timesteps, where n is the average episode duration in timesteps. The graph proves to be the best indicator of the behavior of quadcopter overall episodes on the whole. The graphs in the upper right corner of Fig. 6 and Fig. 7 show the variation of quadcopters location in x, y , z values at each timestep during the episode when the quadcopter earned its highest total reward.

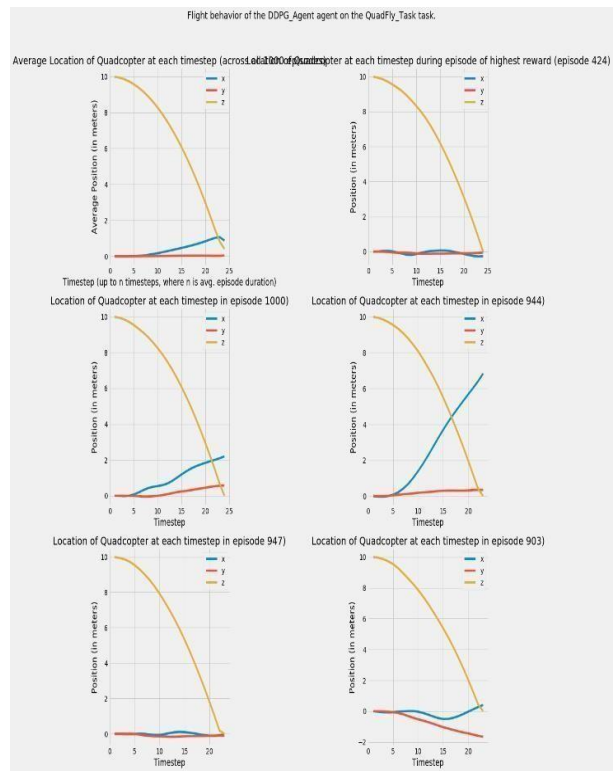

Fig.6. Physical behavior of DDPG agent in simulation

This shows the behavior corresponding to the highest reward in an episode for the quadcopter. The last four graphs in Fig. 6 and Fig. 7 plot x, y, z values of the quadcopter at each timestep for randomly chosen 4 episodes from the final 100 episodes of the simulation, which typically is the period when the two agents have decently and adequately learned the Quadfly task. The graphs show that the agent begins each episode at (0.,0.,10.0) and as the timestep progresses, the agent starts to drift downwards to a height of 0 while also experiencing some drift away from the center (0,0) of the x-y plane. The graph depicts the ARS-agent keeps itself stable at a height of 10, and centered over the center (0,0) of the x-y plane.

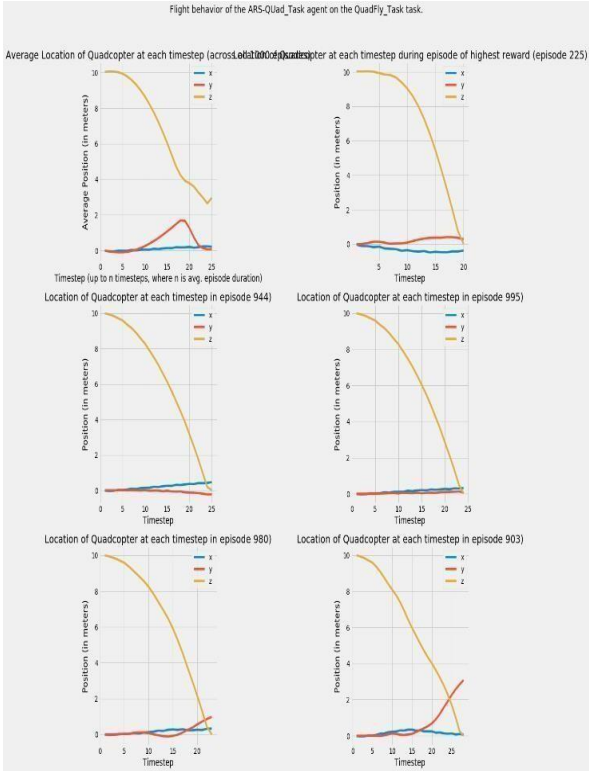

Fig.7. Physical behavior of ARS-Quad agent in simulation

The experimental setup for the simulation purposes involves the initial position $(x_i, y_i, z_i)$ and the target position $(tx_i, ty_i, tz_i)$ for the flying of quadcopter and is set to (0.,0.,10.0) and (0.,0.,150.0) with Euler angles ($\Phi, \theta, \Psi$) as (0.,0.,0.).

The following Fig. 8 and Fig. 9 show the trajectory of the quadcopter from the initial position to the final position during the simulation. They were developed using plotly.

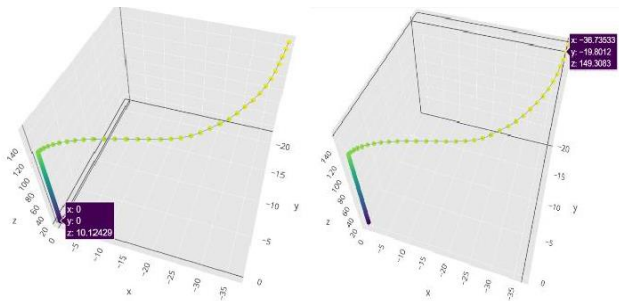

Fig.8. Quadcopter trajectory during simulation

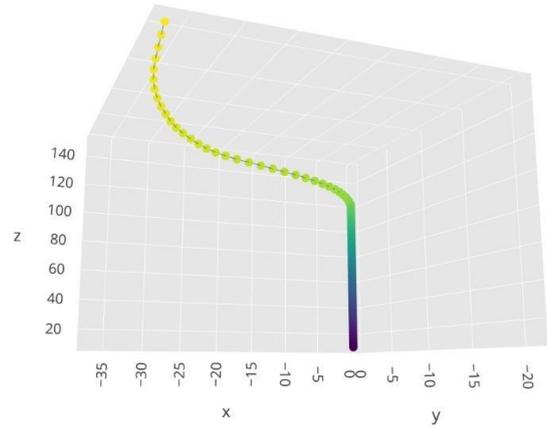

Fig.9. Plotly visualization of Quadcopter motion after stability at 10 m above the ground level

Our simulations dispels the belief of model-based RL always exhibiting superior performance than model-free strategies for continuous control tasks like flying of a quadcopter. The simulations show that ARS-Quad reacts relatively accurately to step response per episode and the agent exhibits significant and better sample complexity in the task of flying. The computational benefits of not making high-level decisions using well trained neural networks are utilized by the ARS-Quad agent to learn the goal relatively accurately and at a faster pace.

## V. CONCLUSION

This research work presented a deterministic policy for the control of quadcopter which is one of the major continuous control problems reinforcement learning is trying to solve. While most of the previous work focuses on solving such problems of autonomous control using well-trained neural networks which require a lot of training data for state-of-the-art results. The present work eliminates this demand of data through a random search in the space of policies and achieves state-of-the-art accuracy for the task of flying the quadcopter. The work demonstrates through simulations how the random search strategy differ in performance from other competitive reinforcement learning strategies for quadcopter control. It indicates that the proposed algorithm is conservative but stable for the task of quadcopter control. The simulations for the DDPG and ARS-Quad agent behavior were carried out for 1000 episodes with randomized seed and noise values and it was observed that the ARS-agent was able to learn the goal relatively accurately and at a faster pace. Also, it was observed that though the ARS-Quad agent had sudden average reward dips for the last 100 episodes unlike DDPG agent, it was continuously able to earn an average reward which was better than that earned by the DDPG agent. Further, the simulations also show the quadcopter flying from the source position to the target position after keeping itself stable at a height of 10m above ground level with a smooth and traceable trajectory. Overall, the proposed searching strategy shows a better sample complexity in the parameter space of policies than the RL strategies that majorly work by exploring the space of actions.

The present work creates abundant room for further progress which can be taken into consideration in future investigations. Future works should consider the establishment of simple baseline methods of random search before moving to complex methods of policy exploration. The present work is a use case of the random search application for autonomous control. In addition, the use of the proposed algorithm for adapting and learning in environments with dynamic properties can be addressed. Another line of future work can be the use of a game-theoretic leader-follower equilibria approach for quadcopter control. The long term goal for the extension of the work can be the comparison of the proposed algorithm with other state-of-the-art RL algorithms for autonomous control.

ACKNOWLEDGMENT

## Authors' Profiles

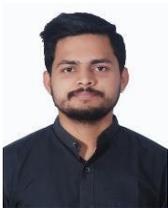

**Ashutosh Kumar Tiwari,** is currrently working as a Member Techincal Staff in Oracle Inc. He has completed his Bachelor of Engineering degree in Information Science and engineering from BMSCE, Banglore. He has interned at R&D labs of two major MNC's, Accenture and Epicor Incorporation and worked on Deep learning and reinforcement learning algorithms to drive business goals. He is an active member of Google developer group Bengaluru. His research interests include ML, DL , Data Engineering and Cloud Computing .

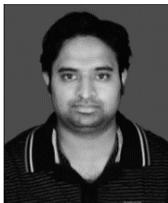

**Sandeep Varma Nadimpalli,** is currently working as Assistant Professor since 2014 in the Department of Information Science and Engineering, B.M.S College of Engineering. He received his B.Tech. degree in Information Technology from JNTU Hyderabad, Telangana, India in 2007. He received his M.Tech. from Andhra University in 2009 and his Ph.D. in computer science and systems engineeringfrom Andhra University in 2015. He also worked as Junior Research Fellow (Professional) from 2010 to 2011 and later worked as Senior Research Fellow from 2011 to 2013 at Andhra University. His research interests include Data engineering, Data Privacy, Cloud Computing and Social Networks. He is a member of IEEE.